\documentclass[sn-mathphys,Numbered]{sn-jnl}% Math and Physical Sciences Reference Style
\usepackage{graphicx}%
\usepackage{multirow}%
\usepackage{amsmath,amssymb,amsfonts}%
\usepackage{amsthm}%
\usepackage{mathrsfs}%
\usepackage[title]{appendix}%
\usepackage{xcolor}%
\usepackage{textcomp}%
\usepackage{manyfoot}%
\usepackage{booktabs}%
\usepackage{algorithm}%
\usepackage{algorithmicx}%
\usepackage{algpseudocode}%
\usepackage{listings}%
\usepackage{indentfirst}
\usepackage[utf8]{inputenc}

\begin{document}

\title[Article Title]{\mbox{EdaCSC: Two Easy Data Augmentation Methods }\\for Chinese Spelling Correction }


\author*[1]{\fnm{Lei} \sur{Sheng}}\email{xuanfeng1992@whut.edu.cn}

\author[2]{\fnm{Shuai-Shuai} \sur{Xu}}\email{sa517432@mail.ustc.edu.cn}


\affil*[1]{\orgdiv{Automated institute}, \orgname{Wuhan University of Technology}, \orgaddress{\street{122 Luoshi Road}, \city{Wuhan}, \postcode{430070}, \state{Hubei}, \country{China}}}

\affil[2]{\orgdiv{School of Software}, \orgname{University of Science and Technology of China}, \orgaddress{\street{No.96, JinZhai Road Baohe District}, \city{Hefei}, \postcode{230026}, \state{Anhui}, \country{China}}}


\abstract{
	Chinese Spelling Correction (CSC) aims to detect and correct spelling errors in Chinese sentences caused by phonetic or visual similarities. While current CSC models integrate pinyin or glyph features and have shown significant progress, they still face challenges when dealing with sentences containing multiple typos and are susceptible to overcorrection in real-world scenarios. In contrast to existing model-centric approaches, we propose two data augmentation methods to address these limitations. Firstly, we augment the dataset by either splitting long sentences into shorter ones or reducing typos in sentences with multiple typos. Subsequently, we employ different training processes to select the optimal model. Experimental evaluations on the SIGHAN benchmarks demonstrate the superiority of our approach over most existing models, achieving state-of-the-art performance on the SIGHAN15 test set. 
}

\keywords{Natural Language Processing, Chinese Spelling Correction, Data Augmentation, Overcorrection}



\maketitle

\section{Introduction}
\label{sec:introduction}

The purpose of the Chinese Spelling Correction(CSC) task is to detect and correct typos in Chinese sentences, which is a sub-task of the grammar error correction task. 
It attracts more and more attention as a precursor task in various natural language processing(NLP) applications, including search query correction\cite{martin-2004-search-engine-queries}, optical character recognition\cite{afli-etal-2016-using} and automatic speech recognition\cite{ERRATTAHI201832}.
Recently, significant progress\cite{zhang-etal-2020-spelling,cheng-etal-2020-spellgcn} has been made in CSC tasks by leveraging pre-trained language models, such as BERT\cite{devlin-etal-2019-bert}.

\begin{figure}
	\begin{center}
		  \includegraphics[width=0.95\textwidth]{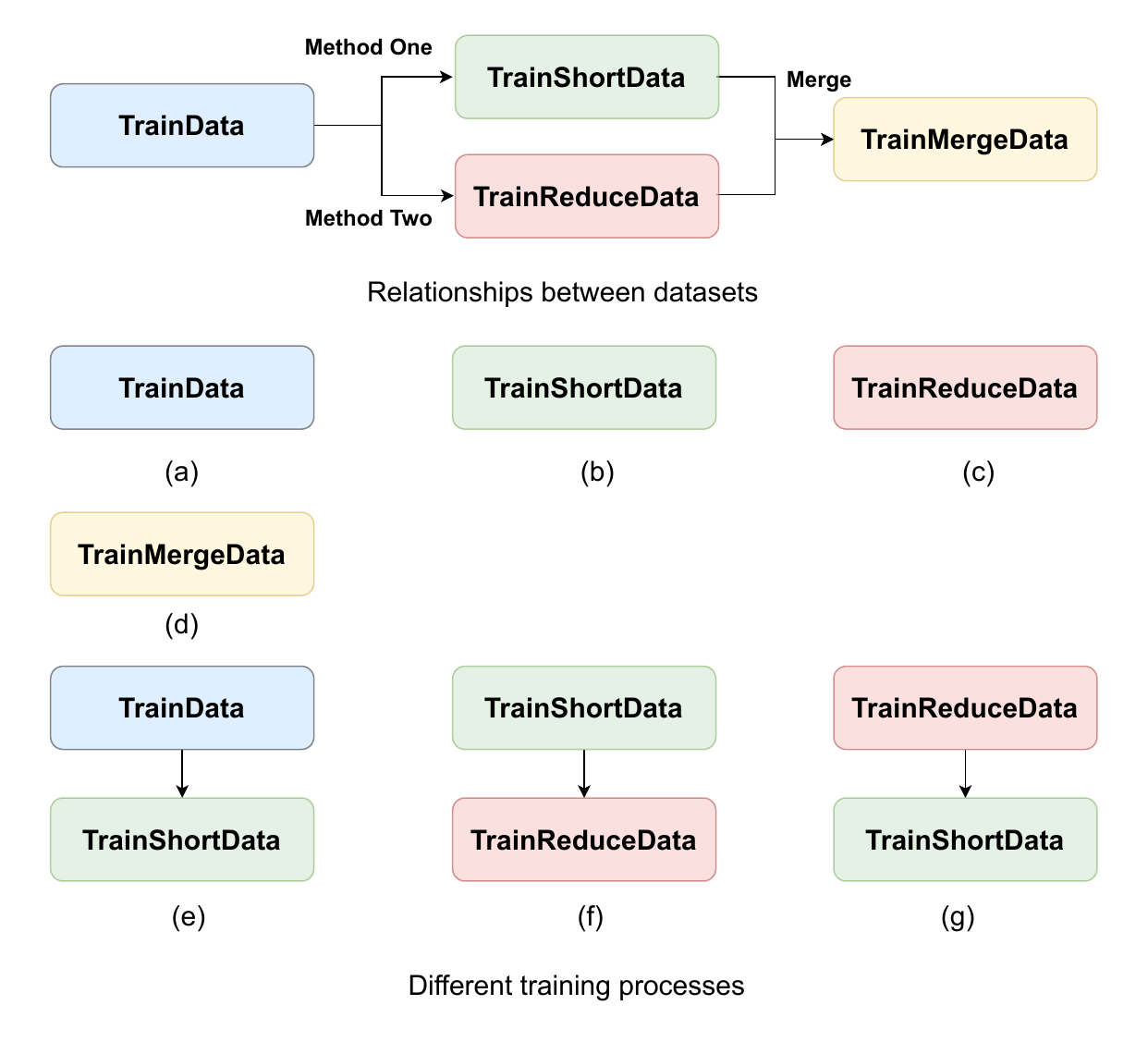}
		  \caption{
		Different types of training datasets and training procedures. The first part shows the relationship between datasets. The original training set \textbf{TrainData} (obtained by the merger of SIGHAN and Wang271K) and two data augmentation methods are used to obtain the the \textbf{TrainShortData} dataset and the \textbf{TrainReduceData} dataset respectively, and then combined to obtain the third dataset \textbf{TrainMergeData}. The following part shows different training processes, a, b, c, and d show that only one dataset is trained alone, e, f, and g show training on the first dataset first, and then the second dataset to train.}
		  \label{figure:eda_dataset} 
	\end{center}
\end{figure}

According to statistics\cite{liu-etal-2010-visually}, Chinese spelling errors are mainly caused by characters with similar shapes or phonetic similarities in Chinese.
Several studies \cite{liu2021plome,xu-etal-2021-read,lv2022general} have enhanced the error correction capability of models by integrating phonetic and glyph similarities into their model. These multimodal features have proven effective in improving the recall ability of the models. However, models that incorporate phonetic and glyph features require pre-training, which necessitates a substantial corpus and computational resources. On the other hand, \cite[]{zhang2022contextual} propose that contextual similarity may hold more value than character similarity for CSC, and enhances the existing CSC model with the Curriculum Learning\cite{cl_survey} approach.
Although much progress has been made, there are still limitations associated with these methods.

\cite{liu-etal-2022-craspell} have observed two limitations in current CSC methods: (1) inadequate performance on sentences with multiple typos; (2) a propensity for overcorrection (originally correct sentence is modified incorrectly), particularly in real-world application scenarios where typos occur infrequently. To address the first issue, \cite{liu-etal-2022-craspell} generate noisy data during the training process. To tackle the second issue, \cite{li-etal-2022-past} propose a loss function based on contrast learning that guides the PLM model to avoid predicting common characters. \cite{gou-etal-2021-think-twice,SHENG2023101573} mitigate the overcorrection problem through the introduction of various post-processing methods. Although these methods have demonstrated improvements, these issues have not been fully resolved.

In this paper, we attempt to address the aforementioned problems by proposing a novel method called EdaCSC, which is short for \textbf{E}asy \textbf{D}ata \textbf{A}ugmentation for \textbf{C}hinese \textbf{S}pelling \textbf{C}orrection. The framework of EdaCSC is illustrated in Figure \ref{figure:eda_dataset}. It comprises two components: (1) augmenting the training dataset using two distinct data augmentation methods, and (2) employing different training procedures for model training. These data augmentation methods are specifically designed to tackle the previously mentioned problems separately. To mitigate overcorrection, we employ the first data augmentation method, which splits long sentences into shorter ones by punctuation marks, thus altering the data distribution and reducing the model's tendency to overcorrect. To handle the degradation in performance for sentences with multiple typos, the second data augmentation method expands the dataset by reducing typos in sentences, enabling the model to learn from less noisy data. After obtaining three datasets by these two data augmentation methods, we finally use different training procedures to select the best model.

Extensive experiments are conducted on the widely used benchmark dataset SIGHAN\citep{wu-etal-2013-chinese,yu-etal-2014-overview,tseng-etal-2015-introduction}. The experimental results demonstrate that our method outperforms all comparative methods on the SIHAN14 and SIHAN15 datasets, and achieves the second-best performance on the SIAHN13 dataset. Furthermore, we conducted ablation experiments on the SIGHAN15 dataset, which reveal that the first data augmentation method effectively improves the precision of the model, reducing overcorrection cases. Similarly, the second data augmentation method proves effective in enhancing the recall of the model, validating the efficacy of our approach.

In summary, our contributions are as follows:
\begin{itemize}
	\item We propose two simple yet effective data augmentation methods that leverage existing datasets to enhance model performance.
	\item We conduct comprehensive experiments and detailed analysis on the SIGHAN benchmark, achieving new state-of-the-art results on SIGHAN15.
\end{itemize}

\section{Related Work}
\label{sec:related_works}
This section introduces the related work of this paper, mainly including Chinese Spelling Correction and Data augmentation.

\subsection{Chinese Spelling Correction}
\label{sec:related_works_csc}
With the development of deep learning techniques, the CSC task has recently achieved great improvements. 
The current mainstream CSC models have leveraged the powerful language modeling capabilities of the BERT \cite[]{zhang-etal-2020-spelling} model and made various improvements. Soft-Masked BERT\cite[]{zhang-etal-2020-spelling} model uses soft-masking technique to connect the error detection network and error correction network. SpellGCN\cite[]{cheng-etal-2020-spellgcn} incorporates phonetic and visual similarity knowledge into language models using graph convolutional networks (GCNs). REALISE\cite[]{xu-etal-2021-read} utilizes a selective fusion mechanism to integrate semantic, phonetic, and glyphic information of Chinese characters, and has become widely adopted as a baseline model due to its effectiveness.

Besides model innovation, other strategies have been explored to enhance error correction in CSC. Two-Ways\cite[]{li-etal-2021-exploration} focuses on improving the generalization and robustness of the CSC model by employing an adversarial attack algorithm to generate adversarial samples for training. ECOPO\cite[]{li-etal-2022-past} observes that pre-trained language models tend to correct wrong characters to semantically correct or commonly used characters. To address this issue, they propose an error-driven contrastive probability optimization method. CL\cite[]{zhang2022contextual} introduces curriculum learning into CSC tasks, which has shown promising results.

\subsection{Data augmentation}
\label{sec:related_works_da}
Data augmentation methods are widely utilized in computer vision(CV)\cite[]{shorten2019survey} and natural language processing(NLP)\cite[]{LI202271}. In CV, techniques like random image flipping and cropping are 
commonly employed, while in NLP, methods such as synonym replacement and back translation are prevalent. For text classification, the easy data augmentation (EDA) method proposed by \cite[]{wei-zou-2019-eda} introduces four operations: synonym replacement, random insertion, random swap, and random deletion. Building upon this, \cite[]{karimi-etal-2021-aeda-easier} further enhances text classification by selectively inserting punctuation marks in the original text, thus preserving word order without introducing excessive noise.

For CSC tasks, data augmentation methods have also been extensively applied to expand datasets.
Previous approaches\cite[]{liu2021plome,wang-etal-2021-dynamic,zhang-etal-2021-correcting,li-etal-2021-exploration} commonly employ confusion sets to generate a large number of training samples through diverse pre-training methods on additionally collected corpora. Fine-tuning is then performed using high-quality training data synthesized by Wang271K\cite[]{wang-etal-2018-hybrid} via ASR and OCR techniques. However, due to variations in the number of pre-trained corpora and construction methods employed, it becomes challenging to conduct fair evaluations and achieve complete result reproducibility. To address this, \cite[]{hu2022cscdime} constructs a new dataset, CSCD-IME, by simulating the Pinyin\footnote{Hanyu Pinyin, often shortened to just pinyin, is the foremost romanization system for Standard Mandarin Chinese. See \url{https://en.wikipedia.org/wiki/Pinyin} for more details.} Input method(IME\footnote{\url{https://en.wikipedia.org/wiki/Input_method}}), and demonstrates the effectiveness of their data augmentation method through comparative experiments. Motivated by these approaches, we propose two simple data augmentation methods that aim to minimize noise when augmenting the dataset.

\begin{figure*}
    \centering
    %\begin{center}
        \includegraphics[width=0.95\textwidth]{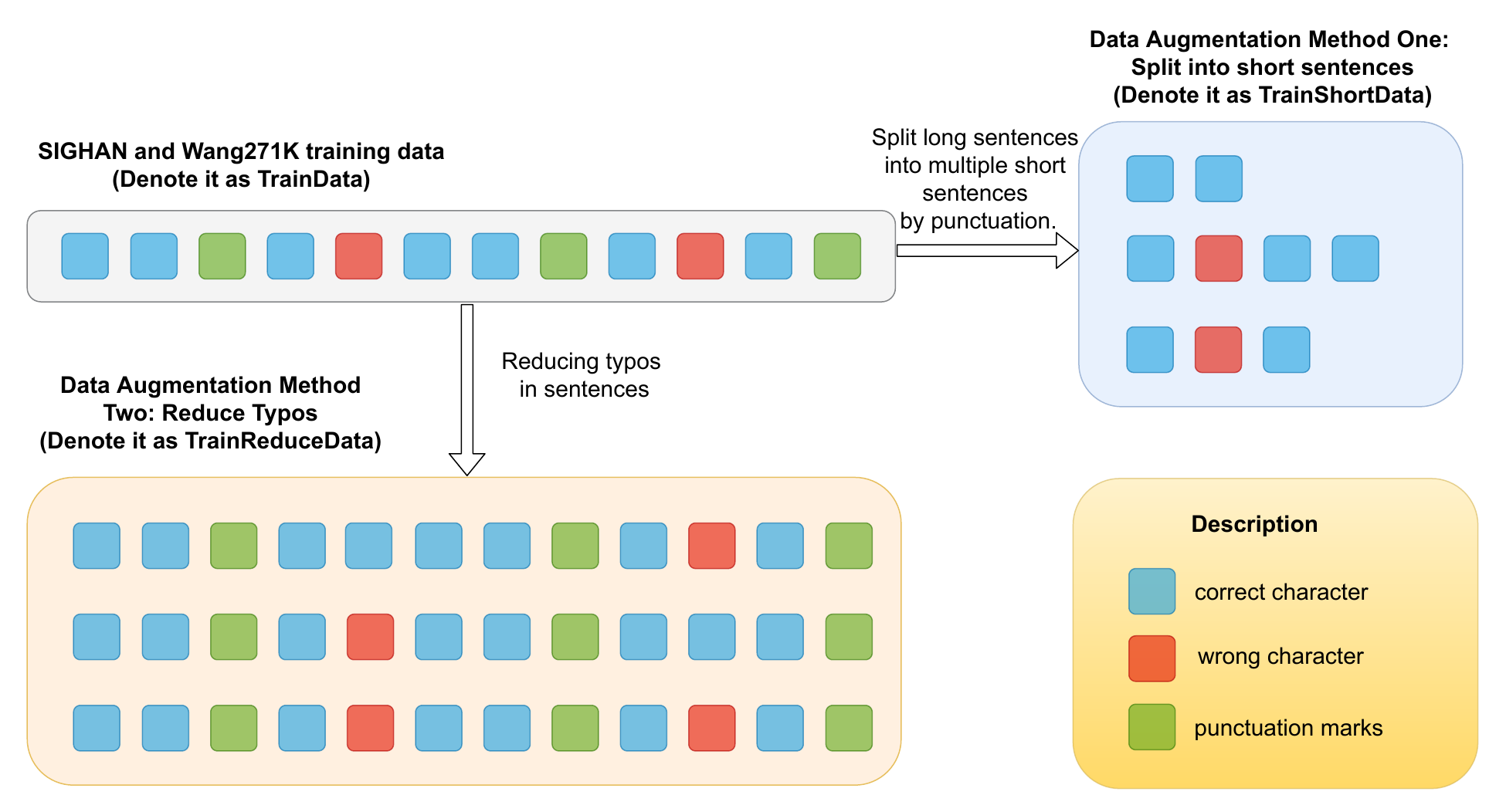}
    %\end{center}
    \caption{Overview of EdaCSC data augmentation methods. We perform data augmentation on the SIGHAN and Wang271K training data by two methods. The first method: split long sentences through punctuation marks (``,", ``.", ``!" , ``?" , ``..." , ``......") as segmentation points into multiple short sentences. The second method: reduce the typos in sentences containing multiple typos in turn, thereby generating multiple sentences. }
    \label{figure:model_csc_eda}
\end{figure*}

\section{Methodology}
\label{sec:methodology}
In this section, we first present the proposed data augmentation method and different training procedures, and then introduce other effective strategies.

\subsection{Data Augmentation}
\label{sec:model_data_augmentation}
The framework of our proposed methods is illustrated in Figure \ref{figure:model_csc_eda}. It mainly includes two data augmentation methods: splitting sentences and reducing typos. 

\textbf{Method One: Split sentences} In order to maintain the semantics of the sentence as much as possible, we select the punctuation marks \{``,", ``.", ``!", ``?", ``...", ``......"\} in the sentence as a cut-off point, thus splitting a long sentence into multiple short sentences. As shown in the right part of Figure \ref{figure:model_csc_eda}: a sentence containing two typos is cut into three short sentences. This idea is inspired by the AEDA\cite{karimi-etal-2021-aeda-easier} method for text classification, which uses punctuation marks randomly inserted into sentences for data augmentation. Intuitively, humans can often correct a sentence with a typo by considering only a part of a short sentence, without depending on the complete sentence. After data augmentation, the distribution of positive and negative samples in the dataset becomes more balanced. We denote the resulting dataset as TrainShortData.

\textbf{Method Two: Reduce Typos} Typos in sentences introduce noise, making sentences with fewer typos easier to correct. In this way, sentences containing multiple typos can be gradually reduced typos for data augmentation. As illustrated in the lower part of Figure \ref{figure:model_csc_eda}: a sentence with two typos is expanded into three sentences through typo reduction (two of the sentences contain only one typo and one original sentence). 98.9\% of the sentences in SIGHAN and Wang171K contain at most three typos, and according to \cite[]{chen-etal-2010-improving}, a sentence contains no more than two typos on average, so we restrict the expanded sentences to contain at most two typos. We denote the resulting dataset as TrainReduceData.

Two new datasets are obtained by the above data augmentation method and then merged to obtain the third dataset,denote as TrainMergeData.

\subsection{Different Training Processes}
\label{sec:different_training_processes}
We obtained three additional datasets by the two data augmentation methods. Considering the distinct characteristics of each dataset, we trained them individually or in combination to determine the best experimental results. The training process is shown in the lower part of Figure \ref{figure:eda_dataset},
with labels 'a,' 'b,' 'c,' and 'd' representing individual training on each dataset, while labels 'e,' 'f,' and 'g' denote combined training. Combined training involves initially training on the first dataset and subsequently retraining with the best model weights on the second dataset.

\subsection{Other Strategies}
\label{sec:other_strategies}
In addition to the aforementioned data augmentation methods, we incorporated several other effective strategies to enhance the model's performance.

\textbf{Pre-trained language models} Pre-trained language models play a crucial role in fine-tuning downstream tasks due to their diverse training corpora and methodologies. ERNIE 3.0\cite[]{sun-2021-ernie3}, which utilizes a large-scale knowledge graph-enhanced corpus for pre-training, has demonstrated state-of-the-art performance in various Chinese NLP tasks. Hence, we adopted ERNIE 3.0 as the underlying pre-trained language model.

\textbf{Further Pre-training} Further pre-training has proven beneficial for CSC, leading to improved results. Unlike many other methods \cite[]{liu2021plome,wang-etal-2021-dynamic,zhang-etal-2021-correcting,li-etal-2021-exploration} that employ confusion sets for constructing pre-training data, we utilized the pseudo-dataset LCSTS-IME-2M\footnote{\url{https://github.com/nghuyong/cscd-ime}}\cite[]{hu2022cscdime}, generated based on the Pinyin input method (IME\footnote{\url{https://en.wikipedia.org/wiki/Input_method}}), for pre-training.

\textbf{Post-processing} \cite[]{liu-etal-2022-craspell} observed that the current CSC model tends to overcorrect, and the error correction effect on sentences containing many typos is poor. To address this issue, SCOPE\cite[]{li-etal-2022-improving-chinese} proposed a simple yet effective constrained iterative correction (CIC) strategy, which effectively mitigates the aforementioned problems. We selected CIC as our post-processing operation.

\section{Experiments}
\label{sec:experiments}
In this section, we first present the details of the experimental implementation and the main results. Then analysis and discussion are performed to illustrate the effectiveness of our method.

\subsection{Datasets}
\label{sec:experiments_dataset}

\begin{itemize}
	\item \textbf{Training Data} Following previous works \citep{cheng-etal-2020-spellgcn,liu2021plome,xu-etal-2021-read}, we use the SIGHAN training data\citep{wu-etal-2013-chinese,yu-etal-2014-overview,tseng-etal-2015-introduction}, which consists of a total of 10K manually annotated data. At the same time, we also included 271K data automatically generated by ASR and OCR techniques, denoted as Wang271K\citep{wang-etal-2018-hybrid}. The SIAHGN and Wang271K are combined and recorded as \textbf{TrainData}. 
	\item \textbf{Eda Data} Our data augmentation is mainly based on \textbf{TrainData}. The dataset enhanced by data augmentation method one is denoted as \textbf{TrainShortData}, and the dataset enhanced by method two is denoted as \textbf{TrainReduceData}. Then, the dataset obtained by merging the \textbf{TrainShortData} dataset and the \textbf{TrainReduceData} dataset is recorded as \textbf{TrainMergeData}. The relationship between these datasets is shown in the upper part of Figure \ref{figure:eda_dataset}.
	\item \textbf{Test Data} We used the SIGHAN test set to evaluate the models, including SIGHAN13, SIGHAN14 and SIGHAN15.
\end{itemize}
	
The detailed statistics of all the data we use in our experiments is presented in Appendix \ref{sec:appendix_dataset}.

\subsection{Baseline Methods}
\label{sec:experiments_method}

To evaluate the performance of our method, we select several latest CSC models as our baselines, include:
\begin{itemize}
	\item \textbf{BERT}\cite[]{devlin-etal-2019-bert}: Directly take the BERT model for fine-tuning on the training set.
	\item \textbf{SpellGCN}\cite[]{cheng-etal-2020-spellgcn}: Fusion of phonological and visual similarities features into BERT models using graph convolutional network(GCN).
	\item \textbf{MLM-phonetics}\cite[]{zhang-etal-2021-correcting}: In this method, uses an end-to-end system based on a pre-trained language model with phonetic features.
	\item \textbf{REALISE}\cite[]{xu-etal-2021-read}: This approach uses the selective fusion mechanism to fuse multiple modal information (semantics, phonetic and graphic) into the model.
	\item \textbf{ECOPO}\cite[]{li-etal-2022-past}: The model uses the idea of contrast learning to allow the model to avoid predicting common characters.
  \item \textbf{LEAD}\cite[]{li-etal-2022-learning-dictionary}: The method enhances the CSC model by learning phonetics, vision, and meaning knowledge from the dictionary.
  \item \textbf{SCOPE}\cite[]{li-etal-2022-improving-chinese}: The model uses auxiliary fine-grained pinyin prediction tasks to enhance CSC model, and proposes a Constrained Iterative Correction post-processing method to alleviate over-correction.
\end{itemize}

\subsection{Experimental Setup}
\label{sec:experiments_setup}

 \begin{table}
	\setlength{\tabcolsep}{1pt}
	\caption{Performance on the SIGHAN13 - 15 test set}
	\label{tab:result_main}
	\begin{tabular*}{\textwidth}{c|l|cccc|cccc}
		\toprule
		\multirow{2 }{*}{Dataset}   & \multirow{2 }{*}{Method}                                                    & \multicolumn{4}{c}{Detection Level} & \multicolumn{4}{|c}{Correction Level}                                                                                                                  \\
		\cmidrule{3-10}
		& & \text{Acc.}                           & \text{Pre.}                            & \text{Rec.}  & \text{F1}     & \text{Acc.} & \text{Pre.} & \text{Rec.} & \text{F1}     \\
		\hline

		%%%%%%%%%%%%%%%%%%%%%%%%%%%%%%%%%%%%%%%%%%%%%%%%%%%%%%%%%%%%%%%%%%%%%%%%%%%%%%%%%%%%%%%%5
		% SIGHAN13
		%%%%%%%%%%%%%%%%%%%%%%%%%%%%%%%%%%%%%%%%%%%%%%%%%%%%%%%%%%%%%%%%%%%%%%%%%%%%%%%%%%%%%%%%5
		
		% SpellGCN 
		\multirow{10}{*}{SIGHAN13} & \text{SpellGCN\cite[]{cheng-etal-2020-spellgcn}} & \text{-}                           & \text{80.1}                            & \text{74.4}  & \text{77.2}    & \text{-} & \text{78.3} & \text{72.7} & \text{75.4}    \\
		% MLM-phonetics  
		& \text{MLM-phonetics$^*$\cite[]{zhang-etal-2021-correcting}}                        & \text{-}                              & \text{82.0}                            & \text{78.3}  & \text{80.1}     & \text{-}    & \text{79.5} & \text{77.0} & \text{78.2}     \\
		% REALISE 
		& \text{REALISE$^*$\cite[]{xu-etal-2021-read}}                 & \text{82.7}                           & \text{88.6}                            & \text{82.5}  & \text{85.4}     & \text{81.4} & \text{ 87.2} & \text{ 81.2} & \text{84.1}    \\
    % ECOPO 
		& \text{ECOPO$^*$\cite[]{li-etal-2022-past}}             & \textbf{83.3}                          & \textbf{89.3}                            & \text{83.2}  & \textbf{86.2}    & \textbf{82.1} & \textbf{88.5} & \text{82.0} & \textbf{85.1}    \\
    % LEDA 
    & \text{LEAD\cite[]{li-etal-2022-learning-dictionary}}             & \text{-}                           & \text{ 88.3}                            & \textbf{83.4}  & \text{85.8}     & \text{-} & \text{87.2} & \textbf{82.4} & \text{84.7}    \\
    % SCOPE 
		& \text{SCOPE$^*$\cite[]{li-etal-2022-improving-chinese}}                 & \text{-}                           & \text{87.4}                            & \textbf{83.4}  & \text{85.4}     & \text{-} & \text{86.3} & \textbf{82.4} & \text{84.3}    \\
		\cmidrule{2-10}
		% BERT compare
		% BERT 
		& \text{BERT\cite[]{devlin-etal-2019-bert}}                & \text{77.0}                           & \text{85.0}                            & \text{77.0}  & \text{80.8}    & \text{75.2} & \text{83.0} & \text{75.2} & \text{78.9}    \\
    % 50041
    & \text{EDA(BERT)}        & \text{82.2} & \text{88.5} & \text{81.9}& \text{85.1}  &\text{81.2} &\text{87.4} &\text{80.8} &\text{84.0}\\

		% 33718		
		& \text{EDA(BERT)$^*$}          & \text{82.8} & \text{88.6} & \text{82.8}& \text{85.6} &\text{81.9} &\text{87.6} &\text{81.9} &\text{84.7}\\
		%%%%%%%%%%%%%%%%%%%%%%%%%%%%%%%%%%%%%%%%%%%%%%%%%%%%%%%%%%%%%%%%%%%%%%%%%%%%%%%%%%%%%%%%5
		% SIGHAN14
		%%%%%%%%%%%%%%%%%%%%%%%%%%%%%%%%%%%%%%%%%%%%%%%%%%%%%%%%%%%%%%%%%%%%%%%%%%%%%%%%%%%%%%%%5
		
		\midrule
		% SpellGCN  
		\multirow{10 }{*}{SIGHAN14} & \text{SpellGCN\cite[]{cheng-etal-2020-spellgcn}} & \text{-}                           & \text{65.1}                            & \text{69.5}  & \text{67.2}    & \text{-} & \text{63.1} & \text{67.2} & \text{65.3}    \\
		% MLM-phonetics  
		& \text{MLM-phonetics$^*$\cite[]{zhang-etal-2021-correcting}}                        & \text{-}                              & \text{66.2}                            & \textbf{73.8}  & \text{69.8}     & \text{-}    & \text{64.2} & \textbf{73.8} & \text{68.7}     \\
		% REALISE  
		& \text{REALISE$^*$\cite[]{xu-etal-2021-read}}    & \text{78.4}                           & \text{67.8}                            & \text{71.5}  & \text{69.6}     & \text{77.7} & \text{ 66.3} & \text{ 70.0} & \text{68.1}    \\
    
		% ECOPO 
		& \text{ECOPO$^*$\cite[]{li-etal-2022-past}}             & \text{79.0 }                          & \text{68.8}                            & \text{72.1}  & \text{70.4}    & \text{78.5} & \text{67.5} & \text{71.0} & \text{69.2}    \\
    % LEDA 
    & \text{LEAD\cite[]{li-etal-2022-learning-dictionary}}             & \text{-}                           & \text{70.7}                            & \text{71.0}  & \text{70.8}     & \text{-} & \text{69.3} & \text{69.6} & \text{69.5}    \\
    % SCOPE 
		& \text{SCOPE$^*$\cite[]{li-etal-2022-improving-chinese}}                 & \text{-}                           & \text{70.1}                            & \text{73.1}  & \text{71.6}     & \text{-} & \text{68.6} & \text{71.5} & \text{70.1}    \\
		\cmidrule{2-10}
		% BERT compare
		% BERT  
		& \text{BERT\cite[]{devlin-etal-2019-bert}}                & \text{75.7}                           & \text{64.5}                            & \text{68.6}  & \text{66.5}    & \text{74.6} & \text{62.4} & \text{66.3} & \text{64.3}    \\
    & \text{EDA(BERT)}           & \text{78.7} & \text{69.9} & \text{70.6}& \text{70.2}   &\text{77.8} &\text{68.0} &\text{68.6} &\text{68.3} \\
		& \text{EDA(BERT)$^*$}           & \textbf{80.2} & \textbf{71.4} & \text{73.1}& \textbf{72.2} &\textbf{79.2} &\textbf{69.4} &\text{71.0} &\textbf{70.2}\\
		%%%%%%%%%%%%%%%%%%%%%%%%%%%%%%%%%%%%%%%%%%%%%%%%%%%%%%%%%%%%%%%%%%%%%%%%%%%%%%%%%%%%%%%%5
		% SIGHAN15
		%%%%%%%%%%%%%%%%%%%%%%%%%%%%%%%%%%%%%%%%%%%%%%%%%%%%%%%%%%%%%%%%%%%%%%%%%%%%%%%%%%%%%%%%5
		\midrule

		% SpellGCN 
		\multirow{10 }{*}{SIGHAN15} & \text{SpellGCN\cite[]{cheng-etal-2020-spellgcn}} & \text{-}                           & \text{74.8}                            & \text{80.7}  & \text{77.7}    & \text{-} & \text{72.1} & \text{77.7} & \text{75.9(74.8)}    \\
		% MLM-phonetics 
		& \text{MLM-phonetics$^*$\cite[]{zhang-etal-2021-correcting}}                        & \text{-}                              & \text{77.5}                            & \text{83.1}  & \text{80.2}     & \text{-}    & \text{74.9} & \text{80.2} & \text{77.5}     \\

		% REALISE
		& \text{REALISE$^*$\cite[]{xu-etal-2021-read}}                 & \text{84.7}                           & \text{77.3}                            & \text{81.3}  & \text{79.3}     & \text{84.0} & \text{75.9} & \text{79.9} & \text{77.8}    \\
    % ECOPO 
		& \text{ECOPO$^*$\cite[]{li-etal-2022-past}}             & \text{85.0 }                          & \text{77.5}                            & \text{82.6}  & \text{80.0}    & \text{84.2} & \text{76.1} & \text{81.2} & \text{78.5}    \\
    % LEDA 
    & \text{LEAD\cite[]{li-etal-2022-learning-dictionary}}             & \text{-}                           & \text{79.2}                            & \text{82.8}  & \text{80.9}     & \text{-} & \text{77.6} & \text{81.2} & \text{79.3}    \\
    % SCOPE 
		& \text{SCOPE$^*$\cite[]{li-etal-2022-improving-chinese}}                 & \text{-}                           & \text{ 81.1}                            & \text{84.3}  & \text{82.7}     & \text{-} & \text{79.2} & \text{82.3} & \text{80.7}    \\
		\cmidrule{2-10}
		% BERT compare
		% BERT
		& \text{BERT\cite[]{devlin-etal-2019-bert}}                & \text{82.4}                           & \text{74.2}                            & \text{78.0}  & \text{76.1}    & \text{81.0} & \text{71.6} & \text{75.3} & \text{73.4}    \\
    & \text{EDA(BERT)}       & \text{86.9} & \text{82.2} & \text{81.7}& \text{81.9} &\text{85.9} &\text{80.1} &\text{79.7} &\text{79.9}\\
    & \text{EDA(BERT)$^*$}      & \textbf{88.0} & \textbf{82.8} & \textbf{84.8}& \textbf{83.8} &\textbf{86.8} &\textbf{80.5} &\textbf{82.4} &\textbf{81.5} \\
		% \cmidrule{2-10}
	
		\bottomrule
	\end{tabular*}

	\footnotetext{ "EDA(BERT)" denotes our proposed method. Best results are in \textbf{bold}. "$^*$" denotes the model is pretrained. For the SpellGCN model, if both precision and recall are correct, the F-score cited in their article should be 74.8.}
\end{table}

% \subsubsection{Evaluation Metrics}
% \label{sec:experiments_setup_metric}

To evaluate the performance of different CSC models, character-level and sentence-level evaluation metrics have been widely used. These include accuracy, precision, recall, F1 score, and false positive rate (denoted as FPR).
 The sentence-level metrics are more stricter than the character-level metrics because there may be multiple typos in a sentence, and a sentence is considered correct only when all typos in a sentence are detected or corrected. We use sentence-level detection and correction metrics.

 Our implementation details are presented in Appendix \ref{sec:experiments_method_detail}.

\subsection{Experimental Results}
\label{sec:experiments_result}

Table \ref{tab:result_main} illustrates the performance of our method and the baseline model on the SIGHAN test set.

From Table \ref{tab:result_main}, we can observe that:
\begin{itemize}
	\item Our EDA(BERT) model outperforms its direct base model BERT across all evaluation metrics and datasets. Specifically, at the correction level, EDA(BERT) achieves a 5.8\% higher F1 score on SIGHAN13, a 5.9\% higher F1 score on SIGHAN14, and an 8.1\% higher F1 score on SIGHAN15.
	Compared with other models, it surpasses all previous state-of-the-art models on the SIGHAN14 and SIGHAN15 test sets, achieves competitive results on the SIGHAN13 test sets. Notably, on the SIGHAN13 dataset, the detection F1 is only 0.6\% lower and the error correction F1 is 0.4\% lower compared to the best model ECOPO\cite[]{li-etal-2022-past}.
	
	\item In comparison to the previous state-of-the-art model SCOPE\cite[]{li-etal-2022-improving-chinese}, EDA(BERT) incorporates further pre-training and employs the CIC post-processing strategy, resulting in higher detection F1 and error correction F1 scores. Particularly, on the SIGHAN15 dataset, the detection F1 is 1.1\% higher and the error correction F1 is 0.8\% higher. This further illustrates the effectiveness of our proposed method.

	\item Notably, our proposed method achieves the highest accuracy and precision rates on the SIGHAN14 and SIGHAN15 test sets, suggesting that it generates significantly fewer instances of over-correction during the error correction process compared to other models. Most models have higher recall than accuracy, and they use fusion of phonetic and glyph features or external knowledge to enhance the error correction capability of the model.
	
\end{itemize}

\begin{table}[htbp]
	\setlength{\tabcolsep}{3pt}
	\caption{ Experimental results on SIGHAN15 test set using different pre-trained language models. }
	\label{tab:result_ablation_plm}

	\begin{tabular}{l|ccc|cccc}

		\toprule
		 \multirow{2 }{*}{Method}   & \multicolumn{3}{c}{Detection Level} & \multicolumn{4}{|c}{Correction Level}      \\
		 \cmidrule{2-8}
		& \text{Pre.}    & \text{Rec.}  & \text{F1}    & \text{Pre.} & \text{Rec.} & \text{F1}   & \text{FPR}  \\
		\midrule
		% EDA(BERT-RoBERTa)  22416
		\text{RoBERTa}   & \text{74.8} & \textbf{79.5}& \text{77.1}  &\text{72.9} &\textbf{77.4} &\text{75.1} &\text{14.3}\\
		% 22318
		\text{+EDA}     & \textbf{79.2} & \text{75.8}& \textbf{77.4} &\textbf{78.0} &\text{74.7} &\textbf{76.3} &\textbf{8.0}\\
		\midrule
		% EDA(BERT-MacBERT)
		\text{MacBERT}    & \text{74.5} & \textbf{77.6}& \text{76.0}  &\text{72.9} &\textbf{76.0} &\text{74.4} &\text{12.2}\\
		% 32093
		\text{+EDA}    & \textbf{79.0} & \text{76.5}& \textbf{77.8}  &\textbf{78.0} &\text{75.6} &\textbf{76.8} &\textbf{7.5}\\
		\midrule
		% EDA(RoCBert) 32184
		\text{RoCBert}     & \text{73.9} & \textbf{78.0}& \text{75.9}  &\text{72.3} &\textbf{76.3} &\text{74.3} &\text{14.1}\\
		% 32082
		\text{+EDA}    & \textbf{80.0} & \text{75.4}& \textbf{77.6}  &\textbf{78.4} &\text{73.9} &\textbf{76.1} &\textbf{7.0}\\
		\midrule
		% EDA(BERT-Ernie3)
		\text{ERINE 3.0}     & \text{77.4} & \textbf{81.9}& \text{79.6}  &\text{75.0} &\textbf{79.3} &\text{77.1} &\text{11.4}\\
		% 18030
		\text{+EDA}        & \textbf{81.5} & \text{79.7}& \textbf{80.6} &\textbf{80} &\text{78.2} &\textbf{79.1} &\textbf{7.5}\\
		\bottomrule
	\end{tabular}
	\footnotetext{The first row of each group denotes the \textbf{TrainData} training set used, "+EDA" denotes the \textbf{TrainShortData} training set used. Best results are in \textbf{bold}. "FPR" denotes the false positive rate, smaller values mean better results.}
\end{table}

\subsection{Analysis and Ablation Study}
\label{sec:experiments_ablation}

We use the BERT base model to perform ablation experiments on the
SIGHAN15 test dataset. 

\subsubsection{Different pretrained language models}
\label{section:difference_plm}
To assess the impact of data augmentation when employing different Pre-trained Language Models (PLMs), We select four PLM (RoBERTa-wwm-ext\footnote{\url{https://github.com/ymcui/Chinese-BERT-wwm}}\cite{cui-etal-2020-revisiting}, MacBERT\footnote{\url{https://github.com/ymcui/MacBERT}}, RoCBert\cite{su-etal-2022-rocbert} and ERINE 3.0\footnote{\url{https://github.com/PaddlePaddle/ERNIE}}\cite{sun-2021-ernie3} ). We conduct comparative experiments using both the \textbf{TrainData} and \textbf{TrainShortData}.

Table \ref{tab:result_ablation_plm} shows the experimental results. The experimental results demonstrate a notable disparity in the correction metrics when different PLMs are employed. Without the EDA method, the maximum difference observed is 2.8\% points, and with the EDA method, it increases to 3.0\% points. This indicates that it is also crucial to choose the appropriate PLM for CSC model. By comparing the detection F1 and correction F1 metrics on all the PLMs after adding the EDA method, the detection F1 and correction F1 metrics were significantly increased, and the correction F1 metrics were improved by 1.8\% points on average, further validating the effectiveness of the EDA method. Due to its superior performance, ERINE 3.0 is chosen as the base model. Additionally, the comparison highlights that our EDA method substantially reduces the FPR by an average of 5.5\% points, indicating its ability to mitigate over-correction.

\begin{table}
	\setlength{\tabcolsep}{1pt}
	\caption{ Experimental results on SIGHAN15 test set using different training datasets.}
	\label{tab:result_ablation_datasets}
	\begin{tabular*}{\textwidth}{l|ccccc|ccccc}
		\toprule
		 \multirow{2 }{*}{Method}   & \multicolumn{5}{c}{Detection Level} & \multicolumn{5}{|c}{Correction Level}      \\
		 \cmidrule{2-11}
		& \text{Acc.}                           & \text{Pre.}                            & \text{Rec.}  & \text{F1}   & \text{FPR}  & \text{Acc.} & \text{Pre.} & \text{Rec.} & \text{F1}   & \text{FPR}  \\

		\midrule

		% TrainData
		\text{(a)TrainData}    & \text{85.3} & \text{77.4} & \text{81.9}& \text{79.6} & \text{11.4} &\text{84.0} &\text{75.0} &\text{79.3} &\text{77.1} &\text{11.4}\\
    % TrainShortData
		\text{(b)TrainShortData}    & \text{86.2} & \textbf{81.5} & \text{79.7}& \text{80.6} & \textbf{7.5} &\text{85.4} &\textbf{80.0} &\text{78.2} &\text{79.1} &\textbf{7.5}\\
    % TrainReduceData
		\text{(c)TrainReduceData}    & \text{85.6} & \text{77.5} & \textbf{82.6}& \text{80.0} & \text{11.4} &\text{84.6} &\text{75.4} &\textbf{80.4} &\text{77.8} &\text{11.4}\\
	% TrainMergeData
		\text{(d)TrainMergeData}   & \text{86.0} & \text{80.6} & \text{81.2}& \text{80.8} & \text{9.3} &\text{85.4} &\text{79.3} &\text{79.8} &\textbf{79.6} &\text{9.3}\\
    % TrainReduceData + TrainShortData
		\text{(g)TrainReduceData + TrainShortData}   & \textbf{86.4} & \text{81.0} & \text{80.6}& \textbf{80.8} & \text{8.0} &\textbf{85.6} &\text{79.4} &\text{78.9} &\text{79.2} &\text{8.0}\\
    \midrule
        %%%%%%%%%%%%%%%%%%%%%%%%%%%%%%%%%%%%%%%%%%%%%%%%%%%%%%%%%%%%%%%%%%%%%%%%%%%%%%%%%%%%%%%%
		% + CIC Post
		%%%%%%%%%%%%%%%%%%%%%%%%%%%%%%%%%%%%%%%%%%%%%%%%%%%%%%%%%%%%%%%%%%%%%%%%%%%%%%%%%%%%%%%%
		\text{+(a)TrainData}            & \text{85.6} & \text{78.3} & \text{82.8}& \text{80.5} & \text{11.6} &\text{84.1} &\text{75.4} &\text{79.7} &\text{77.4} &\text{11.6}\\
		\text{+(b)TrainShortData}             & \text{86.7} & \textbf{82.3} & \text{80.8}& \text{81.5} & \textbf{7.5} &\text{85.8} &\textbf{80.4} &\text{78.9} &\text{79.7} &\textbf{7.5}\\
		\text{+(c)TrainReduceData}             & \textbf{86.9} & \text{79.8} & \textbf{84.7}& \textbf{82.2} & \text{10.9} &\text{85.6} &\text{77.4} &\textbf{82.1} &\text{79.6} &\text{10.9}\\
		\text{+(d)TrainMergeData}   & \text{86.2} & \text{80.9} & \text{81.5}& \text{81.2} & \text{9.3} &\text{85.4} &\text{79.3} &\text{79.8} &\text{79.6} &\text{9.3}\\
		\text{+(g)TrainReduceData + TrainShortData}             & \textbf{86.9} & \text{82.2} & \text{81.7}& \text{81.9} & \text{8.0} &\textbf{85.9} &\text{80.1} &\text{79.7} &\textbf{79.9} &\text{8.0}\\
    \midrule
        %%%%%%%%%%%%%%%%%%%%%%%%%%%%%%%%%%%%%%%%%%%%%%%%%%%%%%%%%%%%%%%%%%%%%%%%%%%%%%%%%%%%%%%%
		% pretrain
		%%%%%%%%%%%%%%%%%%%%%%%%%%%%%%%%%%%%%%%%%%%%%%%%%%%%%%%%%%%%%%%%%%%%%%%%%%%%%%%%%%%%%%%%
    % TrainData
		\text{(a)TrainData$^*$}    & \text{84.6} & \text{76.0} & \text{83.9}& \text{79.8} & \text{14.7} &\text{83.8} &\text{74.5} &\text{82.3} &\text{78.2} &\text{14.7}\\
    % TrainShortData
		\text{(b)TrainShortData$^*$}    & \text{86.7} & \text{81.3} & \text{81.9}& \text{81.6} & \textbf{8.6} &\text{86.0} &\textbf{79.8} &\text{80.4} &\text{80.1} &\textbf{8.6}\\
    % TrainReduceData
		\text{(c)TrainReduceData$^*$}    & \text{85.7} & \text{78.0} & \textbf{84.5}& \text{81.1} & \text{13.1} &\text{84.8} &\text{76.3} &\textbf{82.6} &\text{79.3} &\text{13.1}\\
    % TrainMergeData
		\text{(d)TrainMergeData$^*$}   & \text{86.3} & \text{78.6} & \text{82.8}& \text{80.6} & \text{10.4} &\text{85.6} &\text{77.2} &\text{81.3} &\text{79.2} &\text{10.4}\\
    % TrainData + TrainShortData
		\text{(e)TrainData + TrainShortData$^*$}   & \text{86.6} & \text{80.6} & \text{82.1}& \text{81.3} & \text{9.1} &\text{85.9} &\text{79.3} &\text{80.8} &\text{80.0} &\text{9.1}\\
	% TrainShortData + TrainReduceData
		\text{(f)TrainShortData + TrainReduceData$^*$}   & \text{86.7} & \text{81.6} & \text{82.8}& \text{82.2} & \text{9.5} &\text{85.8} &\textbf{79.8} &\text{81.0} &\textbf{80.4} &\text{9.5}\\
    % TrainReduceData + TrainShortData
		\text{(g)TrainReduceData + TrainShortData$^*$}   & \textbf{87.4} & \textbf{81.7} & \text{83.6}& \textbf{82.6} & \text{8.9} &\textbf{86.3} &\text{79.6} &\text{81.3} &\textbf{80.4} &\text{8.9}\\
    \midrule
       %%%%%%%%%%%%%%%%%%%%%%%%%%%%%%%%%%%%%%%%%%%%%%%%%%%%%%%%%%%%%%%%%%%%%%%%%%%%%%%%%%%%%%%%
		% + CIC Post
		%%%%%%%%%%%%%%%%%%%%%%%%%%%%%%%%%%%%%%%%%%%%%%%%%%%%%%%%%%%%%%%%%%%%%%%%%%%%%%%%%%%%%%%%

		\text{+(a)TrainData$^*$}             & \text{86.4} & \text{79.1} & \text{85.4}& \text{82.1} & \text{12.5} &\text{85.6} &\text{77.4} &\text{83.6} &\text{80.4} &\text{12.5}\\
		\text{+(b)TrainShortData$^*$}             & \text{87.3} & \text{82.2} & \text{83.0}& \text{82.6} & \textbf{8.6} &\text{86.6} &\textbf{80.8} &\text{81.5} &\text{81.1} &\textbf{8.6}\\
		\text{+(c)TrainReduceData$^*$}             & \text{87.1} & \text{80.2} & \textbf{86.3}& \text{83.2} & \text{12.2} &\text{86.2} &\text{78.5} &\textbf{84.5} &\text{81.4} &\text{12.2}\\
    	\text{+(d)TrainMergeData$^*$}       & \text{86.8} & \text{79.7} & \text{84.1}& \text{81.8} & \text{10.6} &\text{85.7} &\text{77.6} &\text{81.9} &\text{79.7} &\text{10.6}\\
		\text{+(e)TrainData + TrainShortData$^*$}  & \text{87.1} & \text{81.5} & \text{83.2}& \text{82.3} & \text{9.1} &\text{86.2} &\text{79.7} &\text{81.3} &\text{80.5} &\text{9.1}\\
		\text{+(f)TrainShortData + TrainReduceData$^*$}   & \text{87.2} & \text{82.4} & \text{83.7}& \text{83.0} & \text{9.5} &\text{86.4} &\text{80.7} &\text{82.1} &\text{81.4} &\text{9.5}\\
		\text{+(g)TrainReduceData + TrainShortData$^*$}            & \textbf{88.0} & \textbf{82.8} & \text{84.8}& \textbf{83.8} & \text{8.9} &\textbf{86.8} &\text{80.5} &\text{82.4} &\textbf{81.5} &\text{8.9}\\
		\bottomrule
	\end{tabular*}
	\footnotetext{ Best results are in \textbf{bold}. "FPR" denotes the false positive rate, smaller values mean better results. Note that "+" means add CIC post processing, and "$^*$" means the model is pretrained.}
\end{table}

\subsubsection{Different training processes}
\label{section:different_training}
In order to fully validate the effectiveness of different data augmentation methods, we conducted comparative experiments using different datasets and training processes, as shown in Figure \ref{figure:eda_dataset}. The experiments results, presented in Table \ref{tab:result_ablation_datasets}, were divided into four groups: the first and second groups did not undergo further pre-training, while the third and fourth groups underwent further pre-training. Additionally, the first and third groups did not employ CIC post-processing, while the second and fourth groups utilized CIC post-processing.

Analyzing the comparison experiments, we can find that:
\begin{itemize}
\item Both further pre-training and adding CIC post-processing can effectively improve the effect of the model.
\item Upon comparing the results of the four cases (a, b, c, and d), it becomes evident that data augmentation method one effectively improves the model's precision, while data augmentation method two enhances its recall. The merged dataset combines the advantages of both methods.
\item Training on one dataset and then on another yields superior results compared to training on a single dataset alone. Specifically, training on the \textbf{TrainReduceData} dataset followed by the \textbf{TrainShortData} dataset yields the best outcomes. 
\end{itemize}

\begin{table}
	\setlength{\tabcolsep}{3pt}
	\caption{ Ablation results on SIGHAN15.}
   \label{tab:result_ablation_study}
	\begin{tabular}{l|ccc|ccc}
		\toprule
		 \multirow{2 }{*}{SIGHAN15}   & \multicolumn{3}{c}{Detection Level} & \multicolumn{3}{|c}{Correction Level}      \\
		 \cmidrule{2-7}
        & \text{Pre.}      & \text{Rec.}  & \text{F1}  & \text{Pre.} & \text{Rec.} & \text{F1}   \\
		
		\midrule
		% EDA(BERT-Ernie3) 
		\text{EDA(BERT)}    & \textbf{82.8} & \text{84.8}& \textbf{83.8} &\text{80.5} &\text{82.4} &\textbf{81.5}\\ 
		\midrule
		% EDA() w/o TrainShortData 
		\text{w/o Short}   & \text{80.2} & \textbf{86.3}& \text{83.2} &\text{78.5} &\textbf{84.5} &\text{81.4}  \\

		% EDA() w/o TrainReduceData 
		\text{w/o Reduce}  & \text{82.2} & \text{83.0}& \text{82.6} &\textbf{80.8} &\text{81.5} &\text{81.1} \\

		% EDA() w/o  FPT
		\text{w/o FPT}   & \text{82.2} & \text{81.7}& \text{81.9} &\text{80.1} &\text{79.7} &\text{79.9} \\
		% w/o CIC 
		\text{w/o CIC}  &  \text{81.7} & \text{83.6}& \text{82.6} &\text{79.6} &\text{81.3} &\text{80.4}  \\
		\bottomrule
	\end{tabular}
	\footnotetext{ The following
	changes are applied to EdaCSC: 
	removing the TrainShortData (w/o Short),
   removing the TrainReduceData (w/o Reduce), 
   removing further pretraining (w/o FPT), 
   removing constrained iterative correction (w/o CIC).
   Best results are in \textbf{bold}. }
\end{table}

\subsubsection{Ablation Study}
\label{section:ablation_study}
The experimental settings that yielded the best results on the SIGHAN15 test set were as follows: employing further pre-training; conducting training on \textbf{TrainReduceData} followed by \textbf{TrainShortData}, and utilizing CIC post-processing method. We performed ablation studies on SIGHAN15 with the following settings:
(1) removing the \textbf{TrainShortData} (w/o Short); 
(2) removing the \textbf{TrainReduceData} (w/o Reduce);
(3) removing further pretraining (w/o FPT);
(4) removing constrained iterative correction post-processing (w/o CIC);
The results are presented in Table \ref{tab:result_ablation_study}.
It is evident that regardless of which component is removed, the performance of EdaCSC declines. This comprehensive demonstrates the effectiveness of each component in our method.

\section{Conclusion}
\label{sec:conclusion}
In this paper, we propose two simple data augmentation methods for Chinese Spelling Correction. The first method involves splitting long sentences into shorter ones using punctuation marks, effectively mitigating overcorrection issues. 
The second method focuses on reducing typos in sentences with multiple typos, thereby enhancing the model's error correction capability. Experimental results demonstrate that our method significantly outperforms most methods and achieves state-of-the-art results on the SIGHAN15 test set. In future work, we will study how to combine other CSC methods to improve the overall error correction ability of the model.

\section{Limitations}
Our proposed data augmentation method demonstrates promising results in Chinese Spelling Correction tasks. However, it requires further validation in other related tasks such as Grammatical Error Correction, Named Entity Recognition, and Text Classification. One limitation of our method is that it cannot be applied to a specific type of dataset where all sentences are short and contain at most one typo. Additionally, it is important to consider that data augmentation increases the dataset size, which in turn extends the model training time.

\section{Declarations}

\textbf{Conflict of interest} The authors have no conflicts of interest to declare that are relevant to the content of this article.

\textbf{Ethics approval} This article has never been submitted to more than one journal for simultaneous consideration. This article is original.

% \textbf{Consent to participate} The authors have approved this article before submission, including the names and order of authors.

% \textbf{Consent for publication} The authors agreed with the content and gave explicit consent to submit.

\textbf{Data Availability} The datasets analysed during the current study are available in the \url{https://github.com/wdimmy/Automatic-Corpus-Generation}. 

\textbf{Code availability} Code and data used in this paper are publicly available at \url{https://github.com/CycloneBoy/csc_eda}.

% \textbf{Authors' contributions}

\begin{appendices}

\section{Statistics of Datasets}
\label{sec:appendix_dataset}

Detailed statistical information of the dataset is shown in Figure \ref{tab:train_dataset}.
It is worth noting that the SIGHAN dataset is in Traditional Chinese. Like previous work, we use the OpenCC\footnote{\url{https://github.com/BYVoid/OpenCC}} tool to convert it into Simplified Chinese.

\begin{table}
	\setlength{\tabcolsep}{3pt}
	\caption{Statistics of the datasets that we use in experiments. }
	\label{tab:train_dataset}
	\begin{tabular}{l|rrr}
		\toprule
		\text{Training Sets} & \text{Sentences}    & \text{Length} & \text{Errors.}   \\
		\midrule
		\text{SIGHAN13}      & \text{350}     & \text{49.2}   & \text{324}                   \\
		\text{SIGHAN14}      & \text{6,528}   & \text{49.6}  & \text{9,699}                \\
		\text{SIGHAN15}      & \text{3,174}   & \text{30.0}   & \text{4,237}             \\
		\text{Wang271K}      & \text{271,329} & \text{42.5}  & \text{381,962}          \\
		\hline
		\text{Total}         & \text{281,381} & \text{42.6}        & \text{396,222}          \\
		\hline
		\midrule
		\text{Eda Datasets} & \text{Sentences}    & \text{Length} & \text{Errors.}   \\
		\hline
		\text{TrainData}    & \text{281,381} & \text{42.6}        & \text{396,222}          \\
		\hline
		\text{TrainShortData}      & \text{724,744} & \text{15.5}  & \text{396,192}          \\
		\text{TrainReduceData}      & \text{546,676} & \text{44.3}  & \text{722,783}          \\
		\text{TrainMergeData}      & \text{1,271,420} & \text{27.9}  & \text{1,118,975}          \\
		\hline
		\midrule
		\text{Test Sets}     & \text{Sentences}    & \text{Length} & \text{Errors.}  \\
		\hline
		\text{SIGHAN13}      & \text{1,000}   & \text{74.3}   & \text{1,224}             \\
		\text{SIGHAN14}      & \text{1,062}   & \text{50.0}   & \text{771}                \\
		\text{SIGHAN15}      & \text{1,100}   & \text{30.6}   & \text{703}                \\
		\bottomrule
	\end{tabular}
	\footnotetext{ \textbf{Length} indicates the average length of sentences in the dataset, and \textbf{Errors} indicates the number of typos contained in the dataset. \textbf{Training Sets} represents the original training set, \textbf{Eda Datasets} represents the dataset in which data augmentation was performed, and \textbf{Test Sets} represents the test set.}
\end{table}

\section{Implementation details}
\label{sec:experiments_method_detail}

Our models are implemented using the Pytorch\cite[]{NEURIPS2019_bdbca288} framework with the Transformer\cite[]{wolf-etal-2020-transformers} library. Our baseline models are BERT\cite[]{devlin-etal-2019-bert} base model, which has 12 transformers layers and 12 attention heads with a hidden state size of 768. We initialize the BERT encoder with the weights of ERNIE 3.0\cite[]{sun-2021-ernie3} base model.

We train our model with the AdamW optimizer for 20 epochs, with learning rate warming up and linear decay, and the warming up step is 10k. The learning rate set 2e-5, a training batch size of 32, and a maximum sentence length of 130.

As mentioned in many previous works\cite[]{cheng-etal-2020-spellgcn,xu-etal-2021-read,li-etal-2022-past}, the mixed use of auxiliaries in the SIGHAN13 test set makes the annotation quality poor. In order to more accurately evaluate the model performance, we used the post-processing method mentioned earlier\cite[]{li-etal-2022-past} and ignored the correction of these auxiliaries.




\end{appendices}


\bibliography{sn-bibliography}% common bib file

\end{document}